%% file: main.tex
\documentclass[letterpaper, 10 pt, conference]{ieeeconf}  

\IEEEoverridecommandlockouts                              

\usepackage{graphics} 
\usepackage{epsfig} 
\usepackage{amsmath} 
\usepackage{amssymb}  
\usepackage{subcaption}
\usepackage{booktabs}

\usepackage{multirow}
\usepackage{verbatim}
\usepackage{amsfonts}    
\usepackage{booktabs}
\usepackage{algorithm}
\usepackage[noend]{algpseudocode} 

\floatname{algorithm}{Procedure}

\usepackage{xcolor}
\usepackage{stfloats}

\usepackage{booktabs} 

\usepackage{xcolor}

\makeatletter
\let\NAT@parse\undefined
\makeatother
\usepackage{hyperref}
\hypersetup{colorlinks,citecolor={blue}} 

\usepackage{graphicx}
\usepackage{etoolbox}
\usepackage{caption}

\title{ARMOR: Egocentric Perception for Humanoid Robot Collision Avoidance and Motion Planning}

\author{
  Daehwa Kim$^{1,*}$ \and
  Mario Srouji$^{2}$ \and
  Chen Chen$^{2}$ \and
  Jian Zhang$^{2}$
  \thanks{\textsuperscript{1}Carnegie Mellon University, daehwak@cs.cmu.edu}
  \thanks{\textsuperscript{2}Apple, \{msrouji, cchen64, jianz\}@apple.com}
  \thanks{\textsuperscript{*}Part of this work was done during an internship at Apple}
}

\makeatletter
\AtBeginDocument{
\let\@oldmaketitle\@maketitle
\renewcommand{\@maketitle}{\@oldmaketitle
  \vspace{3mm}
  \includegraphics[width=\linewidth]{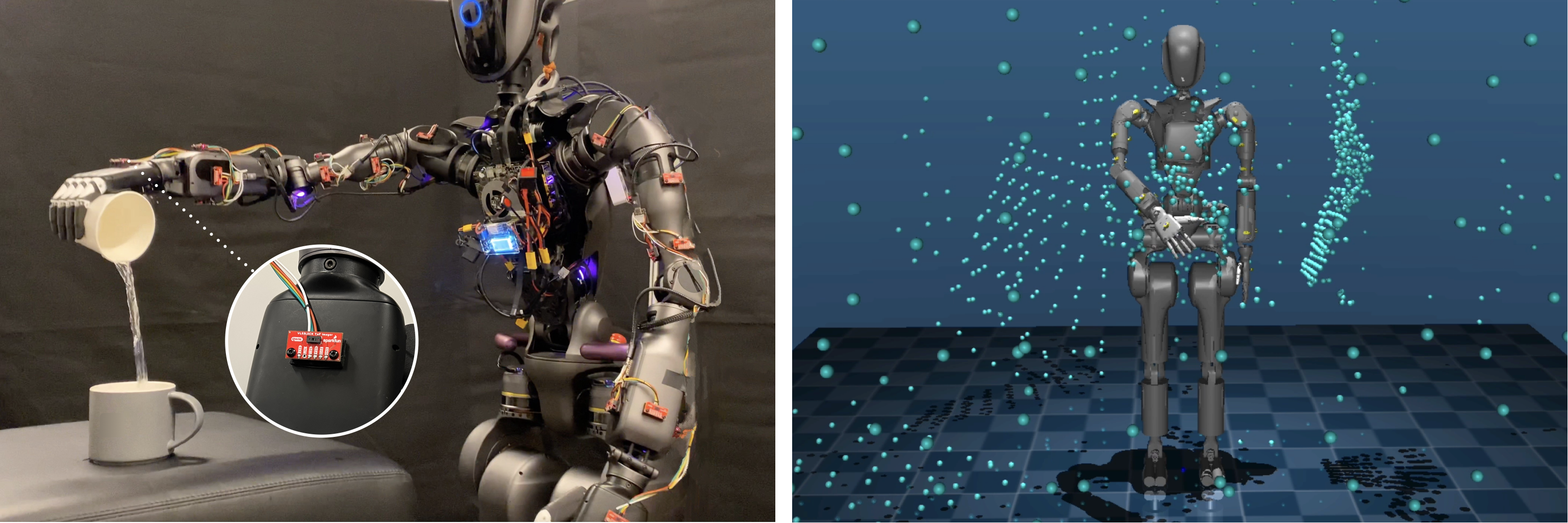}
  \captionof{figure}{
    ARMOR presents a novel egocentric wearable perception hardware and software system for humanoid robots (left). Low-profile and distributed depth sensors enable comprehensive point cloud perception around the robot, and minimize occlusions (right). With a data-driven motion planning policy, ARMOR-Policy, we are able to steer attention to specific regions, and demonstrate effective and fast motion planning.
  }
  \label{fig:teaser}
 }
}
\makeatother

\begin{document}

\maketitle

\thispagestyle{empty}
\pagestyle{empty}

\begin{abstract}

Humanoid robots have significant gaps in their sensing and perception, making it hard to perform motion planning in dense environments. To address this, we introduce ARMOR, a novel egocentric perception system that integrates both hardware and software, specifically incorporating wearable-like depth sensors for humanoid robots. Our distributed perception approach enhances the robot's spatial awareness, and facilitates more agile motion planning. We also train a transformer-based imitation learning (IL) policy in simulation to perform dynamic collision avoidance, by leveraging around 86 hours worth of human realistic motions from the AMASS dataset. We show that our ARMOR perception is superior against a setup with multiple dense head-mounted, and externally mounted depth cameras, with a 63.7\% reduction in collisions, and 78.7\% improvement on success rate. We also compare our IL policy against a sampling-based motion planning expert cuRobo, showing 31.6\% less collisions, 16.9\% higher success rate, and 26$\times$ reduction in computational latency. Lastly, we deploy our ARMOR perception on our real-world GR1 humanoid from Fourier Intelligence. We are going to update the link to the source code, HW description, and 3D CAD files in the arXiv version of this text. 

\end{abstract}

\input{intro}

\input{related}
\input{method}
\input{exp}
\input{conclusion}

{
\small
\bibliography{reference}
\bibliographystyle{plain}
}

\end{document}

%% file: intro.tex
\section{Introduction}

\subsection{Motivation}
The recent advancements in transformer architectures and large language models \cite{bert2021koroteev, radford2018improving, raffel2019t5} have revitalized both interest and applications in humanoid robotics \cite{fu2024humanplus, ze2024humanoid_manipulation}. There are many challenges with humanoid robotic manipulation, including concerns for collision avoidance and safety, many degrees of freedom that make policy learning challenging, and gaps in sensing and perception compared to human skin and tactile feedback. We argue that the current perception and sensing solutions for humanoid robots are not adequate to cover the arms or hands, and do not take advantage of the surface area of the robot.

Due to the mobile nature of humanoid robots, it is not feasible to rely on multiple external cameras and third-person perception like stationary bi-manual manipulators \cite{zhao2023aloha, pmlr-v229-grannen23a, effbiman2020chitnis}. The current paradigm of perception for humanoid robots often involves a centralized camera and/or lidar with high-resolution perception, mounted either in the head or torso \cite{ze2024humanoid_manipulation}. This perception strategy is easy to integrate, and can have decent coverage given a wide field of view, or multiple degrees of freedom neck; however there are still many scenarios where occlusion is present to the arms and hands. Tactile sensing is also integrated into some end effectors \cite{pattabiraman2024visk, haldar2024baku, higuera2024sparsh, lambeta2024digitizingtouch}, however, this drives up the cost significantly, and is hard to integrate in large quantities on humanoid robot arms, and is still not a well-understood source of input in policy learning. 

We present \textbf{ARMOR, a novel egocentric perception for humanoid robotic manipulation and collision avoidance}. We distribute small, low-cost, low-power, Time of Flight (ToF) lidar sensors \cite{sparkfun2024vl53l5cx} across the arms and hand of the humanoid robot strategically to get a good field of view coverage, as well as achieve the desired density of point cloud. These sensors are low-profile and easy to integrate onto humanoid robotic platforms, making them a scalable sensing solution compared to tactile. We also eliminate many occlusions that are present with existing head-mounted or even external cameras. The sensors in our ARMOR perception have also been used in other applications such as mapping \cite{li2022deltardepthestimationlightweight}, collision avoidance \cite{multimodelsensorarray,s24041334}, and 3D reconstruction \cite{nerftof,tof3dreconstruct}.

We use our ARMOR perception to perform collision avoidance on a simulated and real GR1 humanoid robot from Fourier Intelligence. We leverage recent work from ALOHA and others \cite{zhao2023aloha} to learn a dynamic transformer-based motion planner that we call \textbf{ARMOR-Policy}. In order to train our imitation learning policy, we generate ``expert" trajectories by taking 311,922 human realistic motions (86.6 hours worth) from the AMASS \cite{amass2019mahmood} dataset. We then re-target the human arm joints from the data to the Fourier GR1 humanoid robot, and create tight obstacles around the trajectories, resulting in collision-free examples. For better collision avoidance, we also perform a unique inference-time optimization where we sample multiple trajectories. 

Our extensive experiments show that our ARMOR perception is superior when used in both sampling-based and neural-based motion planners. When we deploy our ARMOR-Policy \textbf{with our ARMOR perception, we show a 63.7\% reduction in collisions, and 78.7\% improvement in success rate} when compared with an identical policy using four head-mounted, and externally mounted depth cameras (exocentric perception). We also show that \textbf{our ARMOR-Policy has 31.6\% less collisions, and 16.9\% higher success rate, while being nearly 26$\times$ more computationally efficient} than the sampling-based motion planning expert cuRobo \cite{curobo2023sundar}. Finally we provide numerous qualitative examples of our ARMOR perception, and ACT policy to demonstrate their efficacy in difficult planning scenarios. 

%% file: related.tex
\section{Related Work}

\subsection{Collision-free Motion Planning}
Generating collision-free trajectories for execution is crucial for the safe deployment of general-purpose robotic systems. Humanoid robots, among other systems, are particularly challenging due to their high degrees of freedom in the control space.

Various sample- and optimization-based algorithms have been proposed to tackle collision-free planning problems. These methods often involve a dedicated planner that generates a trajectory, given a collision cost function, either minimizes the cost as part of the optimization objective or prunes and biases the sampling. Popular choices of the collision cost include potential field methods \cite{potentialfield_Flacco2012}, signed distance fields (SDF) \cite{neuromp_Dalal2024, curobo2023sundar, diffseeder_Huang2024a}, and control barrier functions (CBF) \cite{foodprepcbf_Singletary2022, safetycbf_Thirugnanam2022, neurocbf_Yu2024}. 

Other works have also explored learning-based methods through reinforcement learning (RL). The ATACOM method \cite{saferl_Liu2024a, saferlrobo_Liu2022} frames the collision avoidance problem as a manifold optimization problem through constraint manifold theory, and uses it to bound the safety of actions as part of the RL learning. SAFER \cite{safer_Srouji2023}, on the other hand, combines RL with a sample-based algorithm and uses RL to prune the sample space as a runtime optimization to balance task success rate and safety. Our work differs from existing methods by distilling an imitation learning (IL) policy in simulation from an expert motion planner. This allows us to derive a scalable data-driven collision-aware policy that generalizes across different tasks and embodiments (e.g. different robot or sensor configurations).

More recently, training collision avoidance policy has been demonstrated with learning from vast amounts of experience (i.e., imitation learning). M$\pi$Nets \cite{pmlr-v205-fishman23a} and MPNet \cite{neuromp_Dalal2024} train the neural motion planning policy for a robot manipulator arm on several millions of instances using synthetically generated procedural task scenes. Our ARMOR-Policy also employs imitation learning to train the policy in our case study, albeit our perception hardware and data generation pipeline are specifically optimized for humanoid robotic arms. In contrast to other systems that rely on externally mounted cameras, our perception hardware is egocentric and entirely mobile. Additionally, our motion generation is derived from human motion data (e.g., manipulations, social behaviors, dance, etc) rather than task-specific actions as earlier work focused (e.g., pick-and-place operations). Our hope is that this will more closely resemble motions that humanoid robots are likely to do in real-world environments.


\subsection{Egocentric Perception}
The aforementioned collision-aware motion planning methods often require accurate perception of the robot and surrounding objects in a fixed world frame as the input, which necessitates the use of sensors such as an RGB-D camera, or depth sensors mounted external to the robot. 
For humanoid robots with locomotion capability, requiring external sensors could significantly limit the application scenario of the robot, therefore egocentric sensors are usually desirable. 
While some methods can take egocentric sensor inputs \cite{neuromp_Dalal2024}, their collision avoidance performance is often significantly impacted by the limited field-of-view (FoV) of the sensor and occlusions, e.g. head-mounted camera. 
Our work relies on egocentric and distributed ToF sensors that are strategically mounted around the body to form a nearly occlusion-free view of the surrounding environment, thus eliminating the need for any external sensors. 
Recent data-driven bi-manual humanoid robot control policies use egocentric vision as input for policy generation \cite{dphumanoid_Ze2024,okami_Li2024a}. Due to the complexity of humanoid control, most work focuses only on task performance assuming no task-irrelevant obstacles. Using our low-cost, high-coverage, and yet low-dimensional sensor constellation, we envision ARMOR could be integrated into high-level manipulation policies and ensure safe task completion.

\subsection{Use of Time of Flight (ToF) Sensors for Robotic Safety}
There was prior work \cite{multimodelsensorarray, s24041334} that leveraged a similar array of ToF lidar as proximity sensors (single point) for safe Human Robot Interaction, while employing a simple heuristic-based collision avoidance strategy. However, in our work we leverage zone-array-based ToF sensors for more fine-grained and dense obstacle representation and pair this with a more generalizable transformer-based learned policy.

%% file: method.tex
\section{Method}

\subsection{Egocentric Perception Hardware}
\label{sec:hardware}

\begin{figure}
    \centering
    \includegraphics[width=\linewidth]{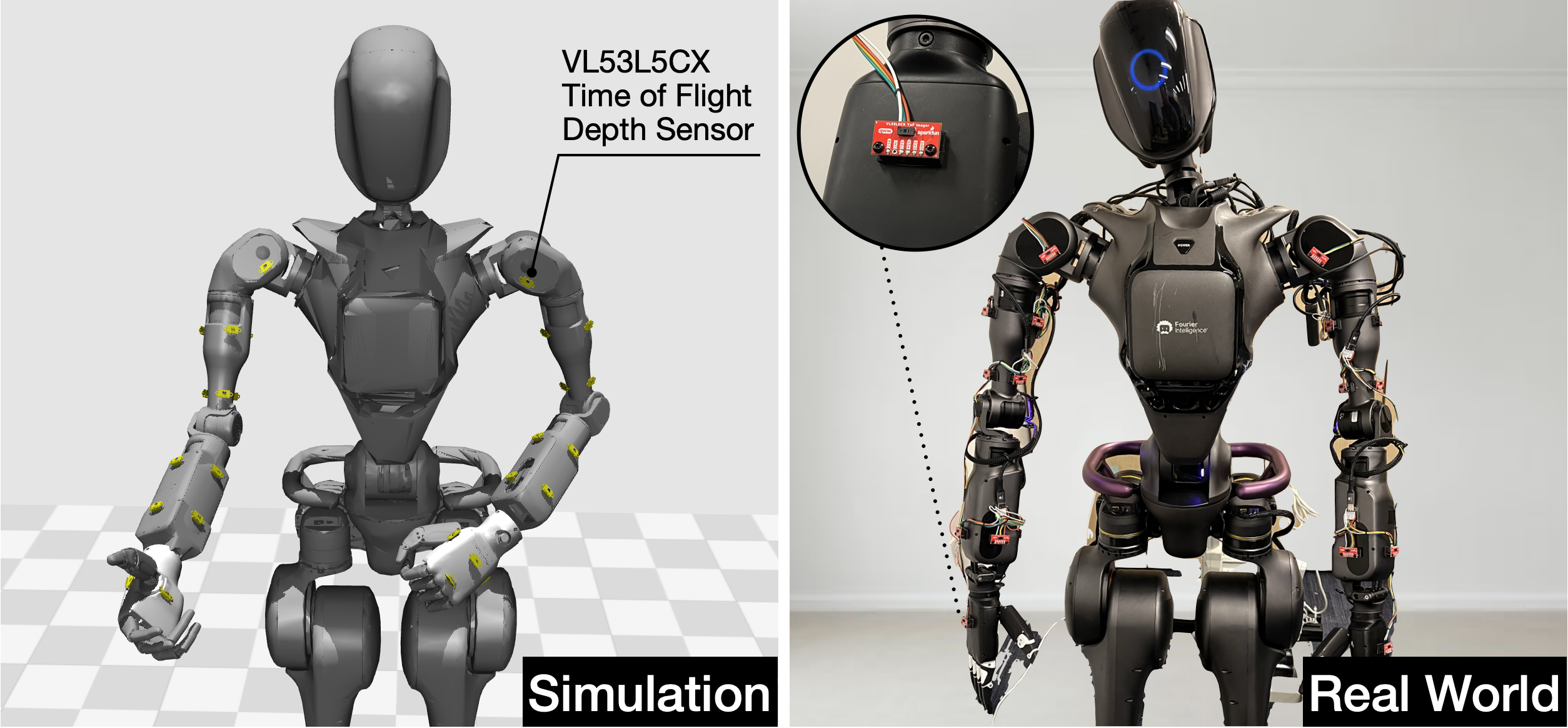}
    \caption{ARMOR's egocentric perception hardware in simulation (left), and deployed on the real robot (right).}
    \label{fig:armor_hardware}
\end{figure}

Unlike centralized RGBD cameras that capture full details in a single dense frame, our approach distributes sparse perception across multiple sensors. This maximally leverages the attention heads of our ARMOR-Policy to attend to different sensor inputs to more effectively plan collision-free trajectories, while being robust to occlusion. 

We chose the SparkFun VL53L5CX time-of-flight (ToF) lidar \cite{sparkfun2024vl53l5cx} for its coarse, yet lightweight, commercially available, and scalable properties. This sensor has a compact dimension of 6.4 $\times$ 3.0 $\times$ 1.5~mm. The sensor runs at 15~Hz (up to 30~hz in certain configurations) with 8 $\times$ 8 resolution of an image, and it captures depth in a 63° diagonal field-of-view and 4000~mm range. Ideally, one can integrate these sensors directly into a humanoid robot's hardware platform, however, for purposes of this work, we attempted to create a solution that can be applied to any humanoid robot using off-the-shelf components.

To demonstrate our sensor constellation, we strategically placed 40 sensors on a Fourier GR1 humanoid robot's arms (20 on each arm - Figure \ref{fig:armor_hardware} left). A group of four sensors is connected to the XIAO ESP microcontroller \cite{xiaoesp32s3}, and is read over the I2C bus. Then, each microcontroller streams over USB to the onboard computer (Jetson Xavier NX) of our robot. Finally, the sensor data is wirelessly streamed over the socket and is processed on a Linux machine with an NVIDIA GeForce RTX 4090 GPU. This ensures the stream can run at 15~Hz even with multiple sensors.

\begin{figure}
    \centering
    \includegraphics[width=\linewidth]{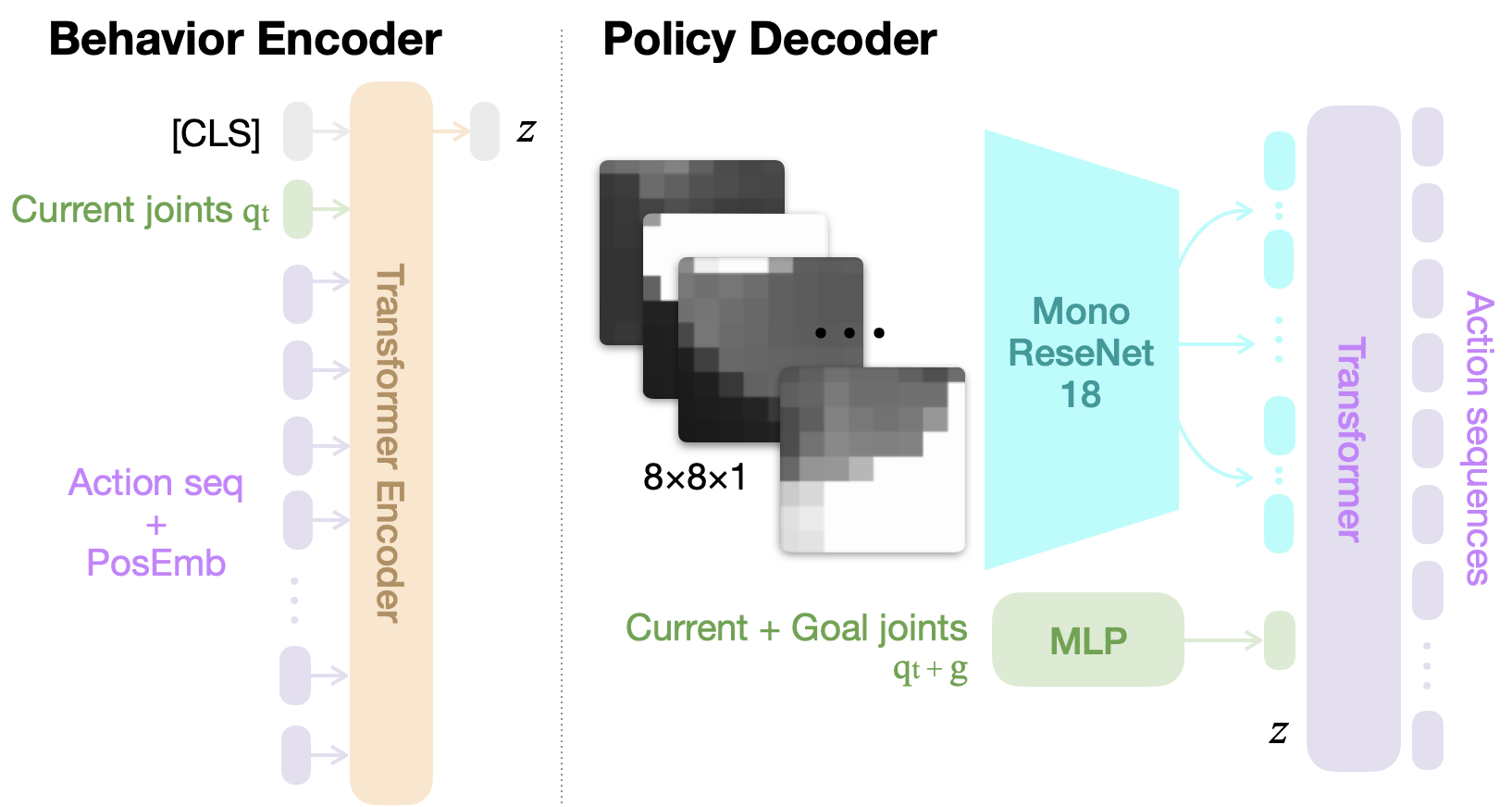}
    \caption{ARMOR-Policy's neural motion planner network architecture. Left: The Behavior Encoder compresses action sequences into style variable $z$, which is later used for diverse output sampling. Right: We implemented the policy decoder to take depth images as input. The depth image is in the lidar camera frame.}
    \label{fig:network}
\end{figure}

\subsection{ARMOR-Policy}

Our ARMOR-Policy is based on a transformer encoder-decoder architecture similar to action chunking transformers (ACT) \cite{zhao2023aloha}, using sequence modeling to imitate the expert (i.e., collision-free human motion demonstrations). We train our policy $\pi(\cdot)$ as a generative model to predict the sequence of actions $a_{t+k}$ conditioned on the current joint state $q_t$, goal joint $g$ positions, observations $o_t$ from the multiple ToF lidar, and latent variable $z$.

Motion planning can yield multiple solutions (e.g., there can be more than one path to avoid obstacles), and our policy should be able to model these behavior sequences. For this reason, we leverage an additional encoder layer to infer the latent variable $z$. The inputs to the encoder are the current joint position and the target action sequences. This encoder is used to train the transformer policy to generate different motion trajectory candidates by adjusting $z$. This flow is illustrated on the left side (`Behavior Encoder') of Figure \ref{fig:network}. This allows us to perform an inference-time optimization where we sample multiple trajectories, as described in section \ref{sec:ito}.

Our policy takes the latent variable $z$, current and goal joint positions, and ToF lidar sensor values as input. The current and goal joints together are a 28-dimensional vector (14 DoF for two arms). We feed each ToF lidar sensor reading in its respective ego-frame into the network. This architecture is shown in the right of Figure \ref{fig:network}. The depth observation includes 40 gray-scale depth images, each with 8 $\times$ 8 resolution. The depth images pass the modified mono-channel ResNet18 backbones \cite{He_2016_CVPR} (with the first layer's weights averaged), which extracts 512 features. Finally, the transformer policy outputs the $k$ action sequences, a $k$ $\times$ 14 vector. This entire architecture yields around 84M parameters.

\subsection{Inference Time Optimization}
\label{sec:ito}
To ensure safe motion plans as in other prior work \cite{neuromp_Dalal2024, pmlr-v205-fishman23a}, we implemented a lightweight inference time optimization. As we mentioned earlier, there can be multiple solutions for collision-free planning, and the ARMOR-Policy is trained to be able to output multiple solutions by adjusting the latent variable $z$. We batch compute $N$ candidate trajectories through the use of the latent style variable $z$, by sampling from random posterior distributions. This step is computed in parallel on the GPU, and adds negligible additional latency during inference. Given multiple output trajectories, we find the optimal path with the least robot-to-point cloud ($PCL$) distance by using the signed distance function ($SDF$). Specifically, the optimization process is formulated as such:


\begin{equation}
\min \sum_{t=1}^{t=T} \sum_{k=1}^{k=K} \{1/SDF_{q_t}(PCL_k)\}
\label{eq:1}
\end{equation}
where $T$ is the action horizon, $K$ is the number of points in the point cloud, and $q$ is the joint position.

\begin{figure*}
    \centering
    \includegraphics[width=\linewidth]{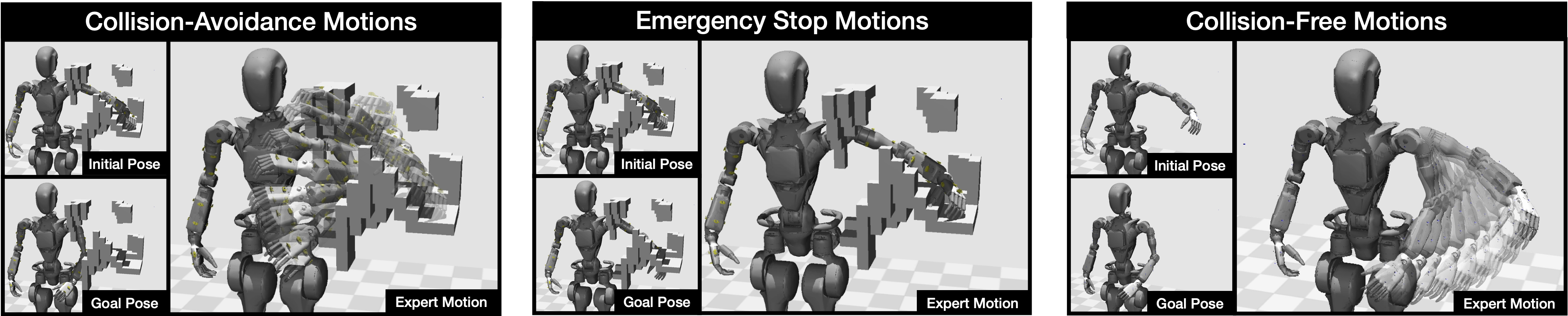}
    \caption{Three data generation strategies. In collision-avoidance motion, a 1-second sequence of human motion in the AMASS dataset is used as a motion planning expert, and the obstacles are placed tightly around, but not colliding with, the motion trajectory. In an emergency stop, the goal pose is randomly chosen inside of the obstacle location. In collision-free motion, we remove all obstacles and linearly interpolate the trajectory from the initial pose to the goal.}
    \label{fig:data_gen}
\end{figure*}

\subsection{Motion Planning Expert Data Generation}
Our core pipeline employs imitation learning. The data includes the current arm pose, a goal arm pose, and environmental obstacles in the form of a simulated point cloud. While other prior work focused on generating motion data in specific task environments (e.g., shelves, cabinets, table, etc) \cite{neuromp_Dalal2024}, our pipeline attempts to learn a set of general manipulation motions around obstacles to avoid over-fitting in specific task environments. We created 311,922 synthetic motion trajectories (86.6 hours worth) using the Archive of Motion Capture as Surface Shapes (AMASS) dataset \cite{amass2019mahmood}, as this data includes diverse human poses that are relevant to robot tasks (e.g., manipulation, dance, social actions, etc). We use these human action trajectories in the AMASS dataset as motion planning expert paths. Just as one might give a narrow environment, and ask the demonstrator to move their arms without colliding, we conversely generate tight obstacles around the re-targeted human trajectory, while ensuring there is no collision with the path. 

We re-target the human arm poses from AMASS to humanoid robot joint configurations specific to the Fourier GR1 humanoid. The AMASS dataset provides a joint angle around each axis in a rotation vector. We use the arm-related joint angles (collar, shoulder, elbow, wrist, and fingers). Each motor of the elbow and wrist in our humanoid robot's arm can use the axis angle directly, as each motor rotates around a single axis. However for the shoulder motors, we use heuristics to combine the angles from both the collar and the shoulder. 

We generate demonstration data via three different strategies as shown in Figure \ref{fig:data_gen}: collision-avoidance, emergency-stop, and collision-free motions. During collision avoidance motions, we use a 1-second future of the current pose in the data sequence as the goal pose, and playback the entire 1-second target action sequence with random obstacles generated around the trajectory, but without actual collision. During stop motions, we set the last goal position to a random location lying inside of an obstacle, which always leads the arm to collide. Lastly, in free motion, the goal pose is the 1-second future, similar to collision avoidance motions, but we removed all obstacles and the expert motion trajectory is linearly interpolated between the current pose and the goal pose.

%% file: exp.tex
\section{Case Study: Collision-Free Motion Planning}

Here as a case study, we evaluate our ARMOR perception, and ARMOR Action Chunking Transformers (ACT) policy as a collision-free bi-manual motion planner for humanoid robots. The experiments aim to evaluate both hardware and software contributions of ARMOR by examining several key aspects: the advantages of ARMOR's egocentric perception approach for collision avoidance and the effectiveness of our ACT policy's neural motion planner architecture for egocentric perception. 

\subsection{Experimental Setup}

\begin{figure}
    \centering
    \includegraphics[width=\linewidth]{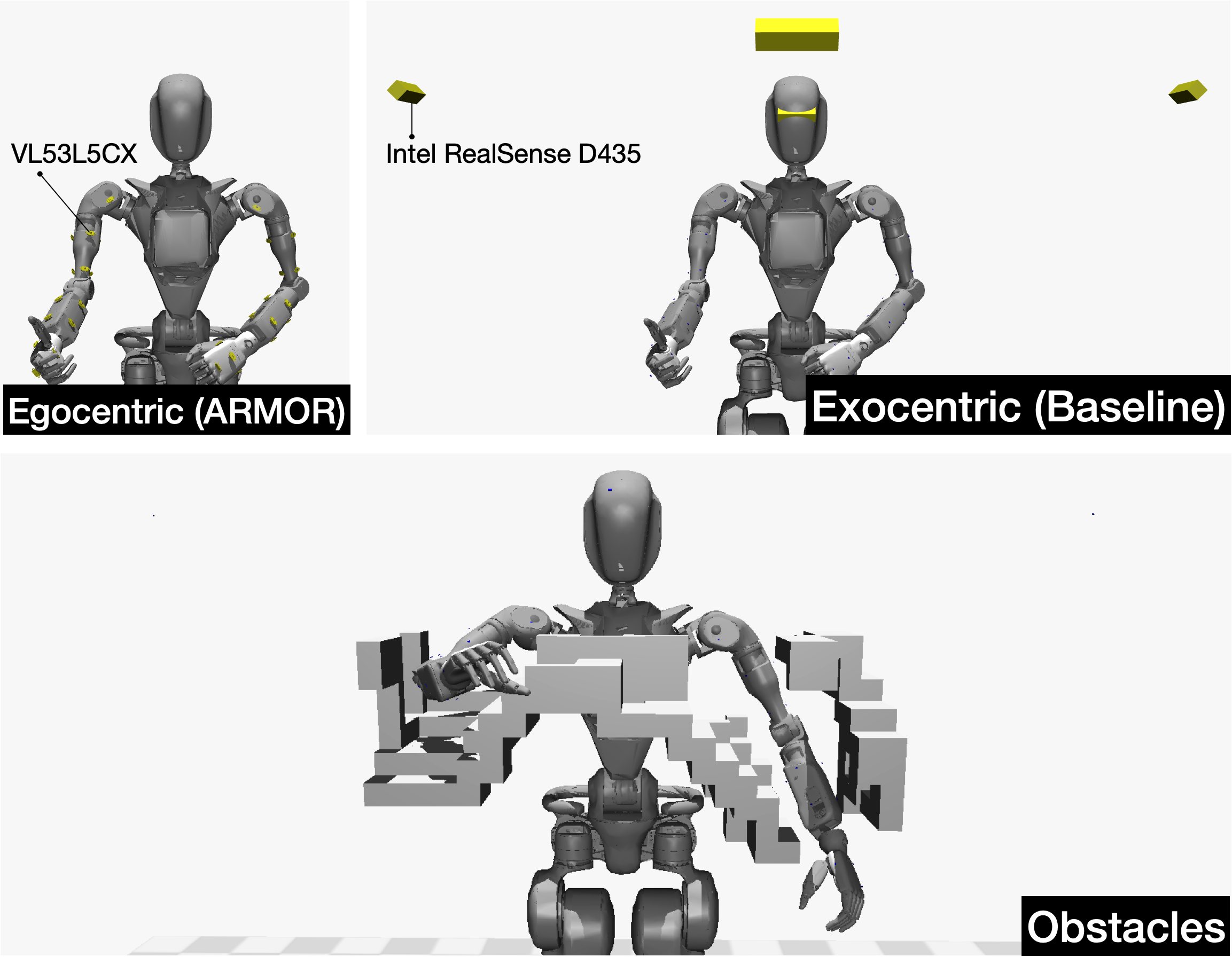}
    \caption{Experiment setup. The yellow geometries indicate depth cameras, Intel RealSense D435 (Exocentric) and VL35L5CX ToF sensor (Egocentric).}
    \label{fig:exp_setup}
\end{figure}

We conduct evaluations in simulation with two different perception configurations: egocentric ARMOR sensor constellations, and external cameras. For egocentric ARMOR sensor constellations, VL53L5CX ToF depth sensors are positioned in 20 different locations on each arm as described in Section \ref{sec:hardware} and Figure \ref{fig:armor_hardware}. For external cameras, we simulate the Intel RealSense D435 due to its popularity in prior work \cite{neuromp_Dalal2024}. This camera has a wider field of view (87° $\times$ 58°), and significantly higher resolution (1280 $\times$ 720) than the VL53L5CX ToF (8 $\times$ 8). Similar to the setup in Neural MP \cite{neuromp_Dalal2024}, four D435 cameras are installed on and around the robot: one on the head and three at head height positioned to the left, right, and front. Hence while we term this camera setup as exocentric, we still included a head-mounted ``egocentric" camera for fair comparison. All cameras are tilted 45° downward to cover the arm's range of motion (Figure \ref{fig:exp_setup} Exocentric). 

We trained our ARMOR ACT policies on 311,922 synthetically generated motions (86.6 hours), and validated them on another 66,840 instances (18.6 hours). In the experimental evaluations below, we only keep the motion sequences where a solution exists, and that have obstacles in the scene (i.e., Collision-Avoidance Motions in Figure \ref{fig:data_gen}). This yields 22,280 motion sequences (6.2 hours) for testing. 

\subsection{Results}

In this section, we compare our ARMOR egocentric perception hardware to the exocentric setup described above, data-driven ARMOR ACT policy (ACT-Depth) to a sampling-based motion planner cuRobo \cite{curobo2023sundar}, and evaluate the benefits of our inference time optimization strategy when combined with the ACT policy (ARMOR-Policy). We use three metrics for comparison: the number of robot-to-obstacle contacts that resulted in a collision, the number of successful sequences that reached the goal pose, and inference computation time. A successful goal pose is defined as the end effector (i.e., hand) reaching within 10 cm of the target position. 

To calculate cuRobo's computation time, we first prune any point clouds that are not reachable by the arm or cause self-collision. This still leaves too many points with external cameras, which makes the cuRobo computation intractable. To resolve this issue, we did additional pruning in denser areas, and kept a maximum of 10,000 points from the original $\sim$3.7M points. In contrast, we did not have these computation issues with our ARMOR ACT policy due to the increased efficiency of using neural-based planning. 

Both the collision and the success metrics are listed as a percentage improvement relative to a baseline, which is described in more detail in the below subsections. Our results will show that the policies that use our ARMOR perception have superior collision avoidance and success rates. In addition, we will show that using neural-based motion planning with our ARMOR ACT policy enjoys superior inference computation time, which is critical for dynamic collision avoidance, and proves to be an effective method for leveraging our ARMOR perception yielding the best combination of results.

\begin{table}[htbp]
    \centering
    \begin{subtable}[t]{0.49\textwidth}
        \centering
        \caption{Egocentric ARMOR v.s. Exocentric}
        \begin{tabular}{lccc} 
            \toprule
            Policy    & Collision           & Success             & Comp. time \\
            \midrule
            cuRobo    & 38.7\% $\downarrow$ & 11.9\% $\uparrow$   & 1300 ms \\
            ACT-Depth & 55.0\% $\downarrow$ & 76.3\% $\uparrow$   & \textbf{50 ms}  \\
            ARMOR-Policy& \textbf{63.7\%} $\downarrow$& \textbf{78.7\%} $\uparrow$   & 240 ms  \\
            \bottomrule
        \end{tabular}
    \end{subtable}
    \hfill
    \begin{subtable}[t]{0.49\textwidth}
        \centering
        \caption{Neural Policy v.s. Sample based}
        \begin{tabular}{lccc} 
            \toprule
            Policy      & Collision  & Success & Comp. time \\
            \midrule
            ACT-Depth\phantom{text}   & 31.6\% $\downarrow$  & 16.9\% $\uparrow$ & 50 ms \\
            \bottomrule
        \end{tabular}
    \end{subtable}
    \hfill
    \begin{subtable}[t]{0.49\textwidth}
        \centering
        \caption{With v.s Without Inference Time Optimization}
        \begin{tabular}{lccc} 
            \toprule
            Policy      & Collision  & Success & Comp. time \\
            \midrule
            ARMOR-Policy   & 19.4\% $\downarrow$  & 1.4\% $\uparrow$ & 240 ms \\
            \bottomrule
        \end{tabular}
        
    \end{subtable}
    \caption{The summary of results. We report the percentage improvement relative to the baseline setup in each of the following: (a) Comparing the ARMOR egocentric perception to the baseline exocentric D435 perception setup. (b) Comparing our ACT policy to the cuRobo motion planner. (c) Comparing the performance improvement by adding our inference time optimization.}
    \label{tab:allresult}
\end{table}

\begin{figure}
    \centering
    \includegraphics[width=\linewidth]{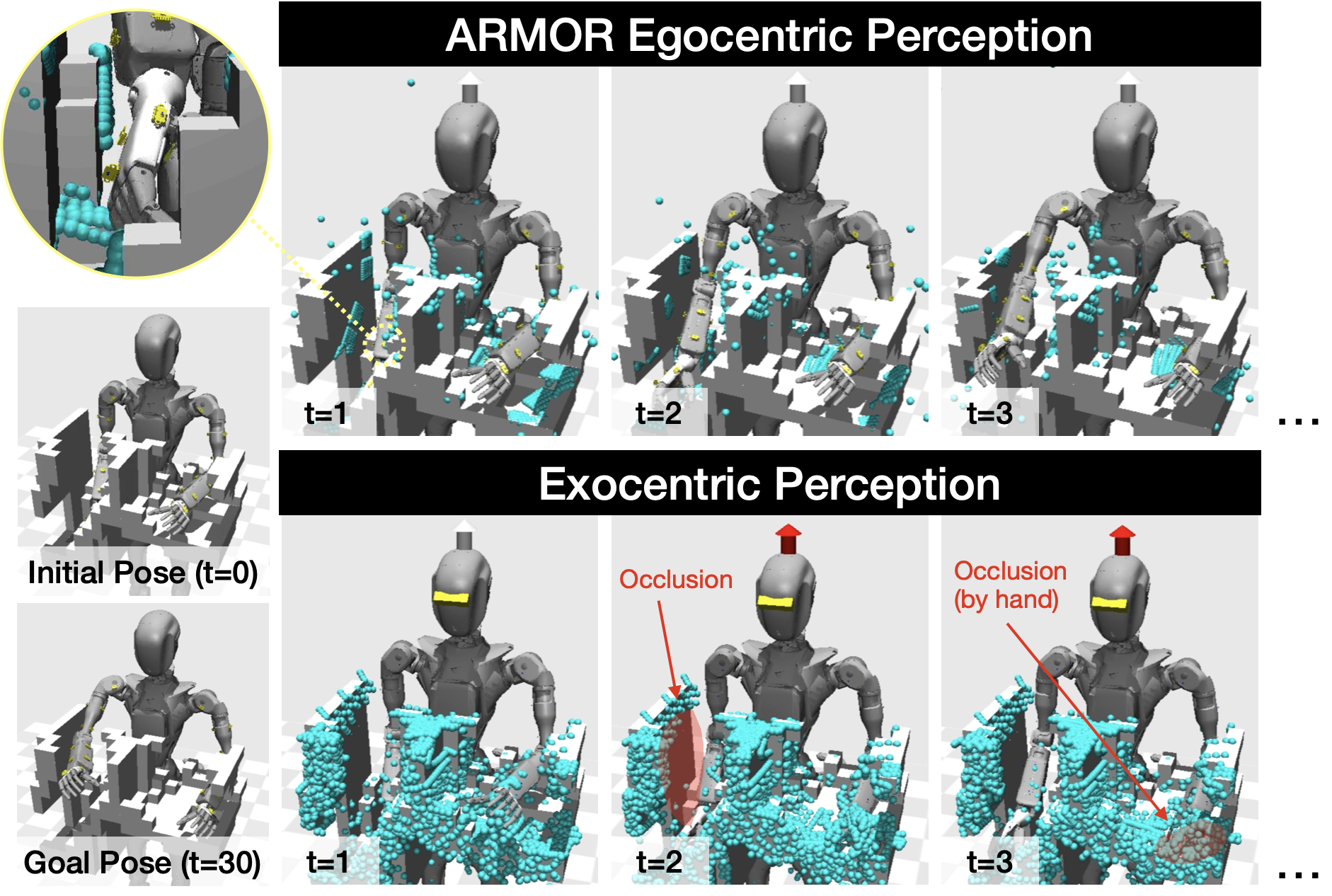}
    \caption{An example of how our ARMOR perception resolves occlusion issues that present in exocentric perception. The visible area from each perception system is visualized as a point cloud in blue. The arrow on the top of the robot head indicates a frame with collision. ARMOR can effectively capture the area around the body (top-left zoomed-in circle) in a cluttered environment. The exocentric perception has a significantly higher resolution, however, it still has many occlusions to both the scene, and the robot's body (bottom red area).}
    \label{fig:nimble_examples}
\end{figure}

\textbf{Egocentric ARMOR v.s. Exocentric Perception.} We compare the performance of ARMOR's egocentric perception system to the exocentric camera setup across three policies: one sampling-based policy (cuRobo \cite{curobo2023sundar}), our data-driven transformer-based \cite{zhao2023aloha} policy (ACT-Depth), and ACT-Depth with our inference time optimization (ARMOR-Policy). Table \ref{tab:allresult} (a) summarizes the performance improvements by percentage relative to the exocentric perception setup for each policy. Across all three policies, ARMOR's egocentric perception outperforms traditional exocentric perception. For example, when our ARMOR-Policy uses ARMOR perception, collisions are reduced by 63.7\%, and success is increased by 78.7\% compared to the exocentric equivalent. Even in our ACT-Depth policy where no inference time optimization is performed, we still see almost similar improvements on both collision avoidance and success. 

Figure \ref{fig:nimble_examples} also illustrates a specific example where exocentric perception can fail inside of a cluttered environment, similar to reaching inside of a drawer or a cabinet. In cases like these, egocentric perception such as ARMOR can allow for more nimble motion planning, especially for high degrees of freedom humanoid robot arms, due to the increased visibility and reduced occlusion where it matters most (alongside the arms).

\textbf{ARMOR ACT Policies v.s. Sampling-based Baseline Policy.} After verifying that our ARMOR perception system outperforms exocentric perception, we compared our transformer-based ARMOR ACT policy to the sampling-based cuRobo baseline. We set the parameters for cuRobo to be the default they describe in \cite{curobo2023sundar}, except we changed the number of waypoints for trajectory optimization to be the same as our transformer model's output. Our testing dataset is quite challenging; it contains tight, and cluttered obstacles around the ground truth motion paths, and we also simulate the sensor noise. Because of this, cuRobo often could not find a solution (for 64\% of the evaluation data). Hence, we only use the sequences where cuRobo was able to find a solution when comparing the collision avoidance results with our ARMOR ACT policy for fairness. 

Table \ref{tab:allresult} (b) shows the performance improvement of our ACT policy without the inference time optimization (ACT-Depth), in comparison to the cuRobo baseline. Both setups are using our ARMOR perception. Our ACT-Depth policy outperforms cuRobo, where collisions are reduced by 31.6\%, and the success instances are increased by 16.9\%. It is also noteworthy that the computation of our ACT-Depth policy is about 26$\times$ faster (50 ms) than cuRobo ($\sim$ 1300 ms). 

\begin{figure}
    \centering
    \includegraphics[width=\linewidth]{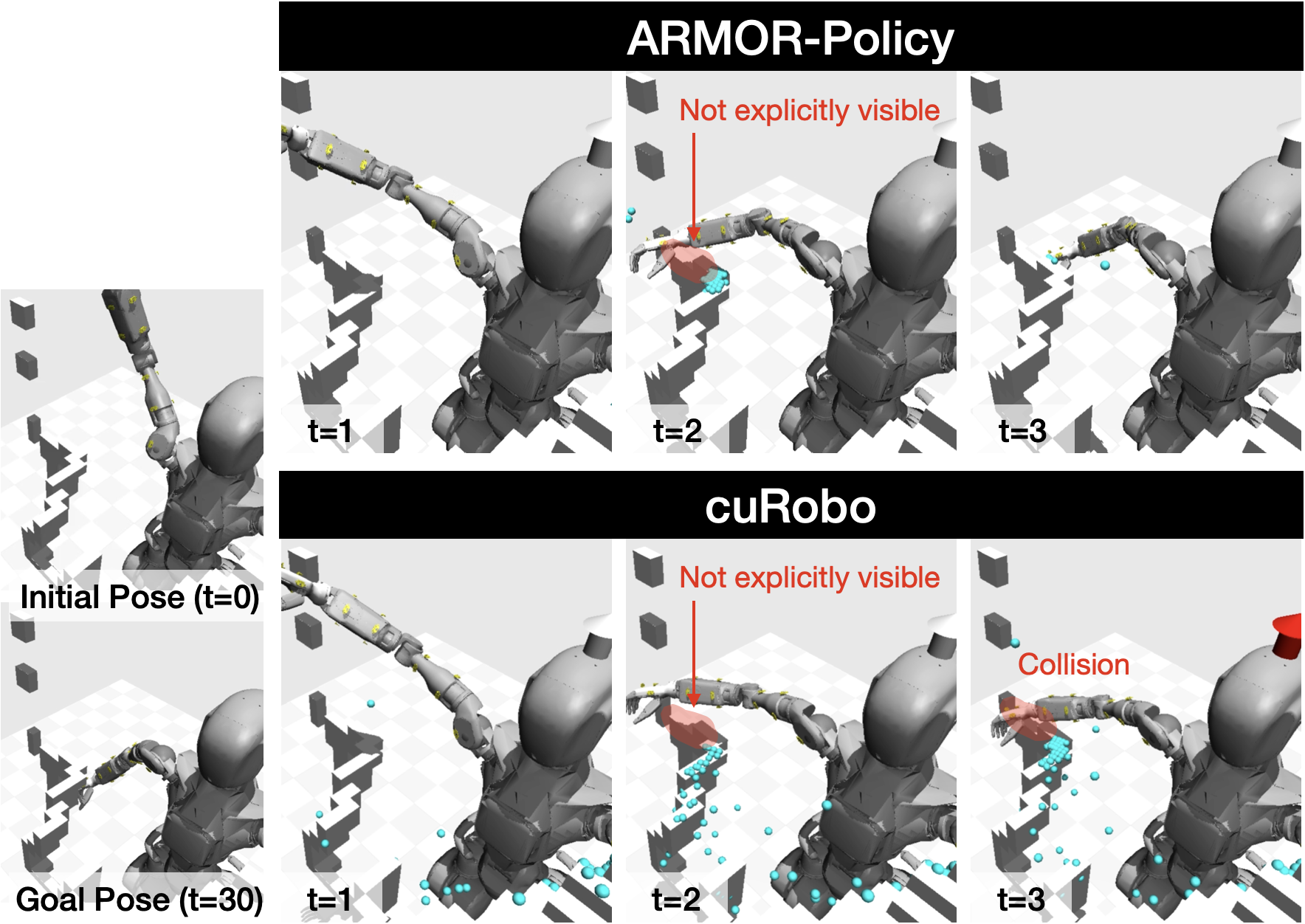}
    \caption{Example of how our ARMOR ACT policy outperforms cuRobo. Both policies use our ARMOR perception. The arrow on the top of the robot head indicates a frame that contains collision. Despite the sparse perception, our ARMOR ACT policy successfully avoids collisions with obstacles (top). The conventional sampling-based planners fail to infer complete obstacle geometry, and hence generate collision-prone paths (bottom).}
    \label{fig:fillgap_examples}
\end{figure}

Figure \ref{fig:fillgap_examples} also illustrates a specific case where our ACT policy outperforms cuRobo. Though our ARMOR perception is sparse, our neural policy is able to effectively infer the gap, and builds proper spatial understanding to avoid the obstacles. However, the sampling-based baseline method accepts the point clouds as-is, and causes collision in the areas that are not explicitly visible.

\textbf{Benefits of Inference-time Optimization.}
As described in Section \ref{sec:ito}, our ARMOR-Policy has a lightweight inference time optimization. Compared to our ACT policy without the inference time optimization (ACT-Depth), our ARMOR-Policy in Table \ref{tab:allresult} (c) shows a further reduction in collisions by 19.4\%, and increases the success rate by 1.4\% compared to ACT-Depth.

\textbf{Real-World Deployment.} We deployed our ARMOR perception with 28 ToF lidars on the Fourier GR1 humanoid. We demonstrated a real-time roll-out of the ARMOR-Policy in a loop for collision avoidance, and update the trajectory at 15-Hz. (ToF lidar update frequency). We will be releasing a link to our ARMOR perception HW+code, ARMOR-Policy code, as well as a video to help in replicating our setup on a real robot in the arXiv version of this text. 

%% file: conclusion.tex
\vspace{-1mm}
\section{Conclusion}
\vspace{-1mm}

We present ARMOR, a novel perception system that leverages wearable sensors for humanoid robotic manipulation to achieve occlusion-free egocentric perception. ARMOR includes both hardware and software contributions. Our low-profile wearable sensor system for humanoid robots enables skin-like distributed receptors and a low-level ARMOR-Policy motion planner for nimble robotic motion trajectories. Our case study showed that ARMOR-Policy can work as a computationally efficient collision-free motion planner in challenging environments. We also demonstrated the effectiveness of egocentric distributed lidar perception in comparison to exocentric cameras. For future work, we hope to leverage our ARMOR perception in dexterous humanoid robotic manipulation tasks, as well as hope that the robotic community continues to expand upon these concepts to further advance the capability of mobile humanoid robots. 

\newpage